\documentclass[runningheads]{llncs}
\usepackage[T1]{fontenc}
\usepackage{graphicx}
 \usepackage{float}
 \usepackage{amsmath}
 \usepackage{algorithm}
\usepackage{algpseudocode}

\begin{document}
\title{Quantifying Similarity: Text-Mining Approaches to Evaluate ChatGPT and Google Bard Content in Relation to BioMedical Literature}
%
%
\author{Jakub Klimczak\inst{1} \and
Ahmed Abdeen Hamed\inst{2}}
\authorrunning{Klimczak, Hamed}
%
\institute{Faculty of Computer Science, AGH, Kawiory 31, 30-055 Cracow, ["jklimczak@student.agh.edu.pl"]\and
Clinical Data Science – Network Medicine and AI, Sano Centre for Computational Medicine
, Nawojki 11, 30-072 Cracow, ["a.hamed@sanoscience.org"]
}
\maketitle              
\begin{abstract}
\textbf{Background -- } The emergence of generative AI tools, empowered by Large Language Models (LLMs),  has shown powerful capabilities in generating content. To date, the assessment of the usefulness of such content, generated by what is known as prompt engineering, has become an interesting research question. \textbf{Objectives -- } Using the mean of prompt engineering, we assess the similarity and closeness of such contents to real literature produced by scientists. \textbf{Methods -- } In this exploratory analysis, (1) we prompt-engineer ChatGPT and Google Bard to generate clinical content to be compared with literature counterparts, (2) we assess the similarities of the contents generated by comparing them with counterparts from biomedical literature. Our approach is to use text-mining approaches to compare documents and associated bigrams and to use network analysis to assess the terms' centrality. \textbf{Results -- } The experiments demonstrated that ChatGPT outperformed Google Bard in cosine document similarity  (38\% to 34\%), Jaccard document similarity (23\% to 19\%), TF-IDF bigram similarity (47\% to 41\%), and term network centrality (degree and closeness). We also found new links that emerged in ChatGPT bigram networks that did not exist in literature bigram networks. \textbf{Conclusions --} The obtained similarity results show that ChatGPT outperformed Google Bard in document similarity, bigrams, and degree and closeness centrality. We also observed that ChatGPT offers linkage to terms that are connected in the literature. Such connections could inspire asking interesting questions and generate new hypotheses.
\end{abstract}

\keywords{Generative AI \and LLM \and Content Assessment \and Google Bard \and ChatGPT \and Text Mining \and Provoking Questions}
%
%
%


\section{Introduction}
\label{introduction}

\begin{figure}[H]
    \begin{minipage}{.43\textwidth}
In November of 2022, our world witnessed an epic event an OpenAI launching pre-trained, and transformer-based Large Language Model, hence its name ChatGPT. The tool is known to be conversational, generative, pre-trained, transformer-based, and hence its name \cite{chatgpt}. The creation of ChatGPT started a new phenomenon in the IT world and implies the appearance of a lot of new models, with various architectures, like Google Bard with PaLM \cite{bard}, \cite{bard2}, \cite{bard3}, or LLaMA \cite{touvron2023llama}. ChatGPT is known for its massive capabilities to receive questions known as prompts in natural languages. It also provides the same human language responses \cite{Moro2023} generated by pre-trained large language models (comprised of massive resources on the web among others). The pre-training takes place by using a transformer-based technology which is implemented using deep-learning methods. They can process extensive amounts of text for tasks such as translation \cite{Mu2023}, \cite{Koraishi2023}, question answering \cite{Guo2023}, \cite{Rao2023} and content generation \cite{Orchard2023}, \cite{Chung2023}, \cite{Agossah2023}. 
    \end{minipage}%
        \hspace{0.02\textwidth}
    \begin{minipage}{.52\textwidth}
\begin{figure}[H]
    \centering
    \includegraphics[scale=.37]{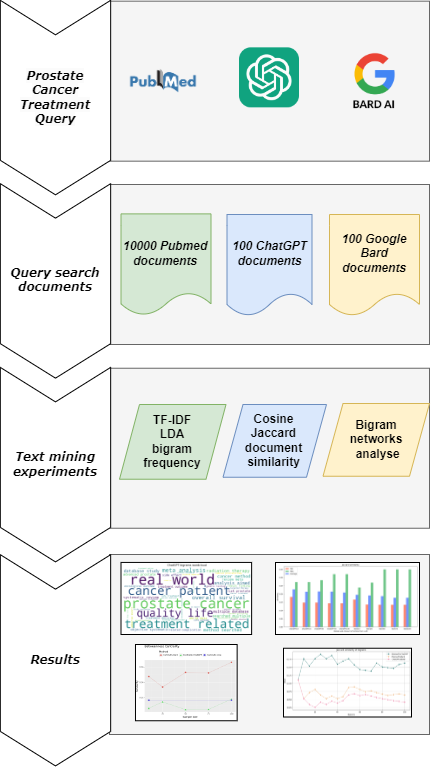}
    \caption{Graphical abstract that demonstrates most important steps of work, from tools that we used, by collecting documents, text mining methods, and results}
    \label{fig:abs}
\end{figure}
    \end{minipage}
\end{figure}

The responses appear to be intelligent and in some cases provide impressive correct answers (in some cases). Generative AI tools are strongly associated with issues such as lack of credibility and transparency \cite{hamed2023challenging}, \cite{hamed2023improving}, \cite{Andre2022} (among many other issues). Such issues in the presence of convincing responses may become irrelevant. As a result, the users of those tools often confuse such responses for truth without questioning and assessing the harm. Nowadays, there are a lot of new AI projects, focused on understandable decision-making. Scientists want to publish their system as fast as possible. Shortliffe in his work claims, that despite the decision making we should focus on the evaluation of constructed solutions \cite{Shortliffe}. Such evaluation has been performed by Roupen et al. \cite{Roupen2024}. where they were validating ChatGPT with oncology questions. Here we are performing the exploratory analysis from an assessment point of view to evaluate the similarity and closeness to real publications. More importantly, we also assess whether there is value in prompt-engineering such tools to generate interesting questions or out-of-the-box hypotheses that could benefit clinicians. In the following sections, we describe in detail the prompt-engineering process for generating content that can be compared with publications and the comparison methods.

Before the launch of generative AI algorithms and LLMs, and the generation of massive data from various resources, Real-World Data (RWD) played a recognizable role in CDSS and diagnostic applications \cite{baumfeld2022current}. Real-world data (RWD) was defined by the US Food and Drug Administration as ``\textit{data regarding the usage, or the potential benefits or risks, of a drug derived from a variety of sources other than traditional clinical trials}`` \cite{fdadocument}. Schad et al. stated that many sources may include patient-generated data including in-home-use data gathered from other sources that can inform on health status (e.g., mobile devices), and data connection approaches \cite{Schad2022}. Bach et al. included PubMed abstracts as another source of RWD \cite{pmlrv70}. Because generative AI tools are pre-trained using Large Language Models (LLMs) \cite{Lee2023a} (including public datasets, crowdsourcing, and customer data) \cite{smithaiblog}, we think that AI generative tools can be conceived as a source of RWD and that it is important, even from an assessment point of view, to investigate their capabilities to conclude whether they may inspire clinicians to ask new and interesting questions and form new hypotheses. it is reasonable to infer that the natural language and conversational capabilities provided by generative AI and LLMs could be transformative to build the next generation of Clinical Decision Support Systems (CDSS). As a result, researchers have explored their potential integration with CDSSs \cite{Liu2023}, \cite{Liao2023}, \cite{Liu2023b}, \cite{Kao2023}.

\section{Data}
We have three different datasets: 

(1) 10,000 \textbf{biomedical literature abstracts}, that are extracted from The PubMed web portal using the search keyword:  ``prostate cancer treatment'' which is our baseline of comparison. Pubmed contains science documents, but also clinical trials and reviews, and all of these types of documents have summary abstracts. We obtain only the first 10000 abstracts of the most frequently cited.

(2) two generative AI datasets of 100 \textbf{abstract-like documents}, received by prompt-engineering ChatGPT and Google Bard. We used these tools to produce documents comparable in structure and form to PubMed abstracts, that is (a random generated ID, a title, and an abstract). To perform this action we used the following query \emph{``Can you generate n reports with titles and abstracts related to "prostate cancer treatment" in a JSON formatted output?''}  which produces a JSON format. LLM tools are based on everything on the internet, so these documents are in abstract format but have sources on the whole internet, not only science 
documents. We have limited the amount of generated articles.

\subsection{Limitation}
At the time of the data collection, we did not have access to the APIs of either tool. We have performed the prompt engineering from the web portals, that was limiting us to generating a maximum of 10 records at a time, we only generated 100 abstracts. Moreover, LLM tools have problems with generating a lot of various data on specific topics and it needs human analysis at the moment. The data is subject to the limitations of ChatGPT and Google Bard. During any generation of the data, we were manually checking if the generated data was related to the topic and if it was formatted correctly. 

\section{Methods}
\label{methods}

We are focused on extracting real medical abstracts from the PubMed portal \cite{pubmed} that will serve as a base to compare with generated data. We also used ChatGPT and Google Bard to generate similar abstracts in the same subject of the search query of PubMed ``prostate cancer treatment''. The abstracts are analyzed using text mining and network analysis. We have performed two text mining methods (1) document similarity using the Cosine and Jaccard similarity\cite{Thada2013}, \cite{Agarwal2014}, (2) bigram frequency and co-occurring comparison with Term Frequency-Inverse Term Frequency (TF-IDF) \cite{Kim2019}, \cite{Wang2019}, \cite{Rani2021}. The network analysis methods are derived from the bigrams forming networks that can be analyzed to compare modularity and term centrality  \cite{Wang2008}, \cite{Meng2021}. 

\subsection{Prompt Engineering}
This work aims to perform prompt engineering in ChatGPT and Google Bard to generate content related to prostate cancer treatment. As remarked earlier, the assumption is that such content may be used in the Clinical Decision Support System (CDSS) and provide links that inspire clinicians to ask unconventional questions. Algorithm \ref{algo:prompt-eng} shows the steps of the prompt engineering process to generate what we call real-world data. Particularly, it is designed to create a set of simulated real-world data reports on prostate cancer treatment. The input parameters include the number of simulated articles (denoted as $n$) and the number of words in each article (denoted as $w$). The algorithm proceeds to generate a list of $n$ reports with titles and abstracts, each abstract consisting of three fields: GPT-ID, Title, and Abstract. The GPT-ID is randomly generated with a maximum length of five alphanumeric characters. Additionally, the abstracts are formatted to have a restricted word count of $m$. The output is provided in a valid JSON format, returned as an array of JSON records, and the overarching topic of investigation is focused on prostate cancer treatment.
\begin{algorithm}
\caption{Prompt-engineering for generating abstract-like documents}
\label{algo:prompt-eng}
\begin{algorithmic}
    \Require The number $n$ of simulated articles.
    \Require The number $w$ of words in each article.
    \State [Content:] Generate a list of $n$ real-world data reports with titles and abstracts.
    \State [Specs:] For each abstract that contains three fields -- GPT-ID, Title, and Abstract -- make it to $m$ words.
    \State [Specs:] Make the GPT-ID random, containing at most five letters and numbers.
    \State [Format:] A valid JSON format returned as an array of valid JSON records.
    \State [Topic:] Investigating prostate cancer treatment.
\end{algorithmic}
\end{algorithm}

\subsection{Text Mining Similarity Analysis}
Following the step of generating the reports, we further compare the content of both tools using traditional text mining. This includes: (1) comparing the documents against other documents, (2) extracting and comparing bigrams of words, and (3) constructing networks of bigrams with identifying novel links. We conducted a comparison using random samples of [10, 25, 50, 75, 100] generated documents against the Pubmed document corpus. for the \textbf{Document Similarity Analysis} -- we count the similarity between real literature medical abstracts and the reports that were generated using ChatGPT and Google Bard. We are looking for the most matching pairs of generated-real articles. Assessing the similarity of documents as units is likely to lead to examining the utility of the generated data. For this task, we used two different metrics: (1) the Cosine similarity \cite{Thada2013}, and the Jaccard similarity \cite{Agarwal2014}. On the other hand, in \textbf{Bigram  Analysis} a Bigram term is a sequence of two next elements extracted from a string of tokens, usually composed of letters, syllables, or words \cite{ngramint}. For example, the following words are bigrams in a document related to prostate cancer: ('prostate cancer', 'cancer cells', 'treatment option', 'risk factors'). This section will measure the frequency similarity of bigrams extracted from literature and documents generated by generative AI tools. Bigrams can be used for the creation of word graphs, that will offer a model that can be used to explore topology and structural property \cite{Liu2019a}. Here, we use the Term Frequency-Inverse Document Frequency (TF-IDF) \cite{tfidf} method with the different datasets of bigrams to count the importance of bigrams within the documents. 






\subsection{Networks Analysis}
Bigrams have a natural possibility to construct interesting networks that can be analyzed for their topology and structural properties. To demonstrate, we will consider the following bigrams: ``prostate cancer'', ``cancer cells'', ``enlarged prostate'', ``prostate health'', ``cancer disease'', ``cancer research''. Connecting the bigrams words using the bigram relationship, offers connectivity. More importantly, the common works also act as a linking node to connect more than one bigram. For instance, the prostate cancer bigram, being comprised of prostate and cancer, offers two individual network nodes: prostate and cancer. If we incrementally add a new bigram (e.g., ``cancer disease'') to the network, only one node is added since the ``cancer'' already exists. This makes the current network comprised of three nodes ``prostate'', ``cancer'', and ``disease'', and two edges (prostate, cancer) and (cancer disease) respectively. 

\begin{figure}[H]
    \begin{minipage}{.38\textwidth}
The continuation of adding new bigrams will result in a network that offers insights into structural properties. Figure ~\ref{fig:exgraph} shows the idea of creating bigram networks using a small number of bigrams. It is now possible to inspect this network from various points of view: (1) the connected vs disconnected components, that is three connected components, (2) the largest connected components (i.e., cancer and prostate), (3) central components (vertices with the highest degree, also cancer and prostate). Such insights may cause interesting questions when various networks are compared. 
    \end{minipage}%
        \hspace{0.02\textwidth}
    \begin{minipage}{.57\textwidth}
\begin{figure}[H]
    \includegraphics[width=.98\linewidth]{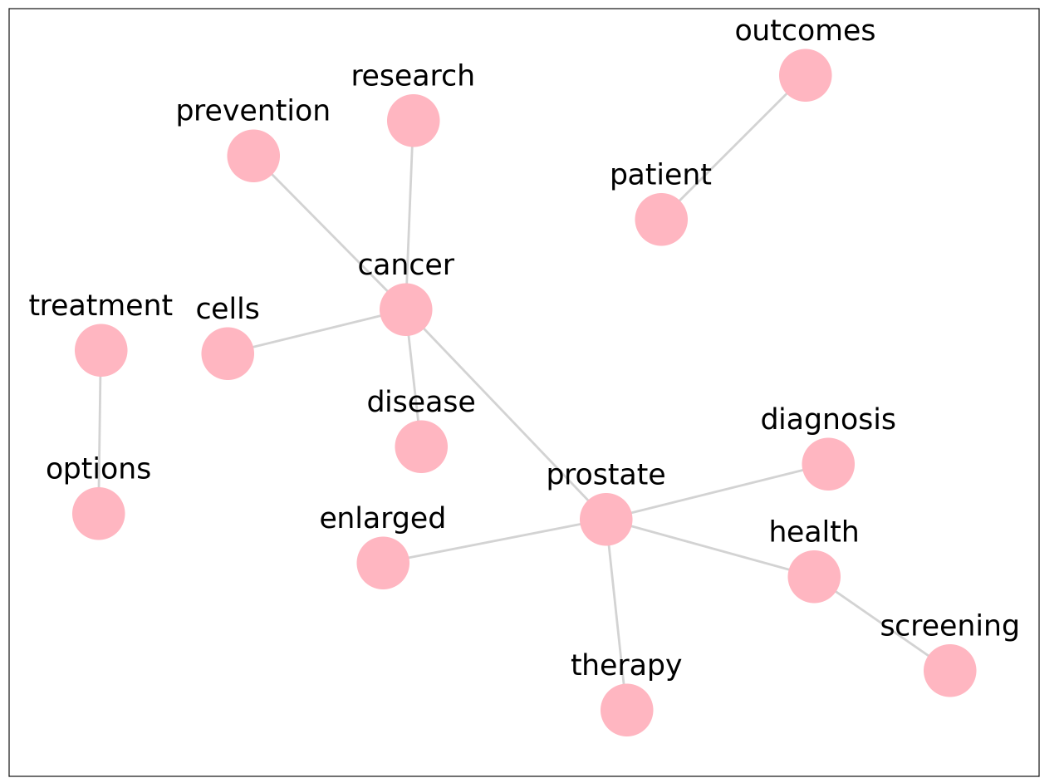}
    \caption{Example graph of bigrams terms}
    \label{fig:exgraph}
\end{figure}
    \end{minipage}
\end{figure}

These approaches enable us to dissect the structural relationships in different terms in the generated and real data. 
Bigram networks are a popular approach for text analysis \cite{hamed2015measuring}, \cite{abdeen2021fighting}, \cite{hamed2023improving}. Such networks can be used in various tasks, including text classification, sentiment analysis, pattern recognition in text, and topic modeling.
To manifest this idea, we select the top 50 most frequent bigrams from the entire corpus of documents from PubMed, ChatGPT, and Google Bard and build the bigram networks as rigorous models of comparison. The type of analysis we present is degree and closeness centrality as common measures that demonstrate the differences. The degree presents how many connections a specific unigram (word) has in the graph \cite{Zhang2017}. A higher value of degree centrality can demonstrate that a bigram or word is significant in terms of connectivity with other words in the network. 

The closeness presents how close a given word is to all others in the same graph \cite{Zhang2017}. The smaller the values, the closer the word is to all other words. By comparing centrality metrics across different data sources, we can get valuable knowledge about similarities and differences in the texts. 

\section{Results}


\subsection{Document Similarity Analysis}
We performed various document similarity experiments to measure the similarity scores of various dataset sizes from  Google Bard and ChatGPT with PubMed article abstracts. We performed two different types of experiments to assess the similarities : (1) using the Cosine similarity, and (2) using the Jaccard similarity. To ensure unbiased experiments, we performed 5 experiments of different dataset sizes (10, 25, 50, 75, and 100). Each of the datasets was selected at random. Table ~\ref{tab:doc-sim} shows the result of the scores calculated for both similarity metrics. While the Cosine similarities (on average) showed similarities that it $>$ 90\% the Jaccard similarities fluctuated from 17\% to 21\% (on average).

\begin{figure}[h]
    \begin{minipage}{.45\textwidth}
        \begin{table}[H]
        \caption{Document Similarity Results: Average Similarity Scores}
        \label{tab:doc-sim}
        \begin{tabular}{cccccc}
        \hline
        {\textbf{Sample Size}} & {\textbf{Source}} & \textbf{Cosine} & \textbf{Jaccard} \\
        \cline{3-4}\cline{5-6}%
        & & \textbf{Score} & \textbf{Score} \\
        \hline
        \textit{10} & \textit{Bard} & 0.3435 & 0.1954 \\
        \cline{2-6}
        & \textit{ChatGPT} & 0.3802 & 0.2287 \\
        \hline
        \textit{25} & \textit{Bard} & 0.3389 & 0.1914 \\
        \cline{2-6}
        & \textit{ChatGPT} & 0.3604 & 0.2139 \\
        \hline
        \textit{50} & \textit{Bard} & 0.3336 & 0.1873 \\
        \cline{2-6}
        & \textit{ChatGPT} & 0.3612 & 0.2157 \\
        \hline
        \textit{75} & \textit{Bard} & 0.3205 & 0.1775 \\
        \cline{2-6}
        & \textit{ChatGPT} & 0.3595 & 0.2147 \\
        \hline
        \textit{100} & \textit{Bard} & 0.3202 & 0.1775 \\
        \cline{2-6}
        & \textit{ChatGPT} & 0.3531 & 0.2093 \\
        \hline
        \end{tabular}
        \end{table}
    \end{minipage}%
        \hspace{0.06\textwidth}
    \begin{minipage}{.35\textwidth}
        \begin{table}[H]
            \caption{Bigram Analysis Results: Average Scores}
            \label{tab:bigram_analysis_results}
            \begin{tabular}{cccc}
            \hline
            {\textbf{Sample Size}} & \textbf{Source} & \multicolumn{2}{c}{\textbf{TF-IDF}} \\
            \cline{3-4}
            & & \textbf{Average} \\
            \hline
            \textit{10} & \textit{Bard} & 0.3999 \\
            \cline{2-4}
            & \textit{ChatGPT} & 0.4182 \\
            \hline
            \textit{25} & \textit{Bard} & 0.40473 \\
            \cline{2-4}
            & \textit{ChatGPT} & 0.44100 \\
            \hline
            \textit{50} & \textit{Bard} & 0.37391 \\
            \cline{2-4}
            & \textit{ChatGPT} & 0.46699 \\
            \hline
            \textit{75} & \textit{Bard} & 0.3814 \\
            \cline{2-4}
            & \textit{ChatGPT} & 0.4556 \\
            \hline
            \textit{100} & \textit{Bard} & 0.36766 \\
            \cline{2-4}
            & \textit{ChatGPT} & 0.42217 \\
            \hline
            \end{tabular}
        \end{table}
    \end{minipage}
\end{figure}

\subsubsection{Cosine Document Similarity Analysis}
Results in table \ref{tab:doc-sim} show that the average similarity scores for this method are very high, between 32-38 \% for both solutions. ChatGPT shows more similarity than Bard. ChatGPT's similarity is around 35-38 \%, and Bard's average similarity is around 32-34 \%. In the plot \ref{cosineimg}) we can see the trend, that for every sample, the average similarity score is higher for ChatGPT. Analysis of table \ref{tab:doc-sim} shows values for 50 documents, where the similarity of ChatGPT is above 36 \%, and of Google Bard is 33 \%.   

\subsubsection{Jaccard Document Similarity Analysis}
\begin{figure}[H]
    \hspace{0.03\textwidth}
    \begin{minipage}{.45\textwidth}
        \begin{figure}[H]
        \includegraphics[width=\textwidth]{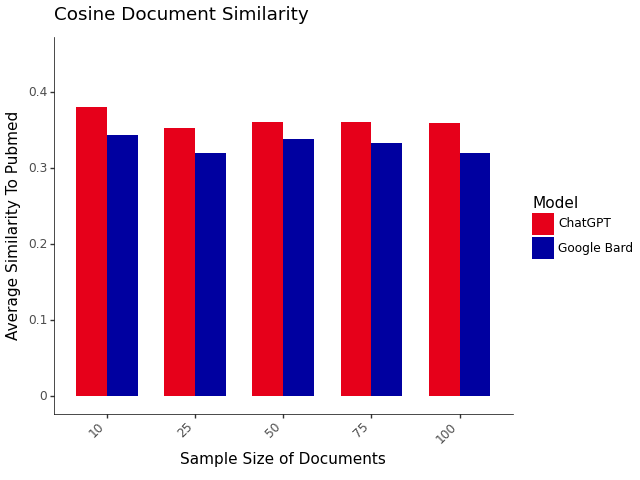}
        \caption{Cosine similarity of documents from PubMed and generative AI tools documents\label{cosineimg}}
        \end{figure}   
    \end{minipage}
        \hspace{0.03\textwidth}
    \begin{minipage}{.45\textwidth}
        \begin{figure}[H]
        \centering
        \includegraphics[width=\textwidth]{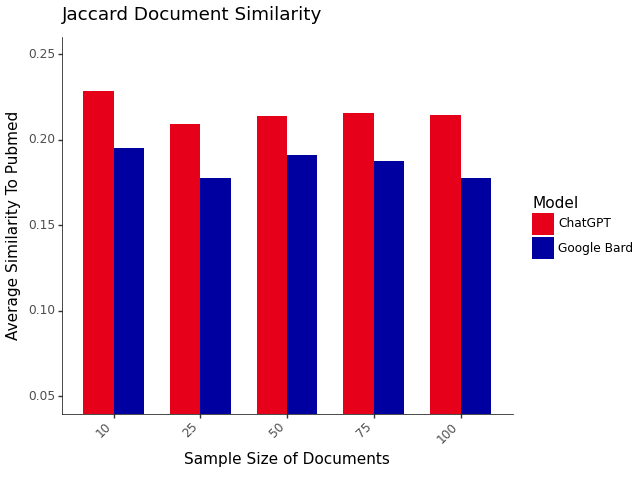}
        \caption{Jaccard similarity of documents from PubMed and generative AI tools documents\label{jacimg}}
        \end{figure}   
    \end{minipage}%
\end{figure}

The Jaccard similarity analysis is performed with a word-bag representation of text, that counts the number of common words by the whole set of words. As a result, the obtained similarity can seem to be not so high, but we need to remember, that the generated texts are shorter than PubMed ones. The average scores (table \ref{tab:doc-sim} are hovering around 18-23 \%. This result is considered good and indicates a good connection between the generated data and real-world data. ChatGPT shows more similarity in average scores, than Bard. ChatGPT's similarity is around 21-23 \%, and Bard's average similarity is 17-19 \%. Plot \ref{jacimg}) demonstrates the advantage of ChatGPT one more time, like in the Cosine case. Table \ref{tab:doc-sim} demonstrates average values. For example, 50 documents similarity of ChatGPT is almost 21.5 \%, and Google Bard is 18.7 \%. It is quite a big difference.
        
\subsection{Bigram Similarity -- TF-IDF Bigram Frequency Analysis}
\begin{figure}[H]
\centering
\includegraphics[width=0.6\textwidth]{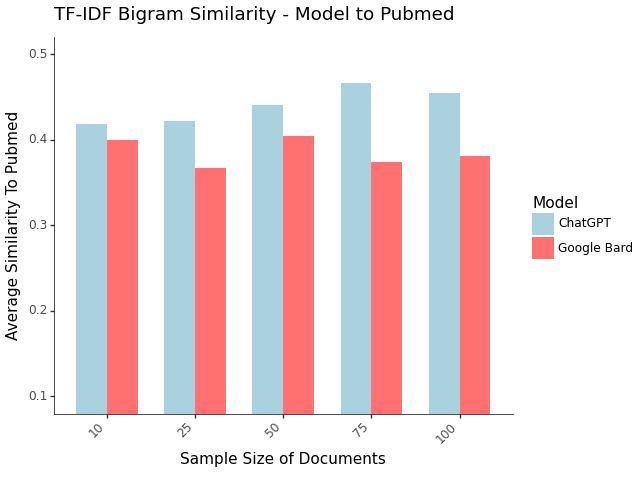}
\caption{TF-IDF similarity of bigrams from PubMed and generative AI tools documents: ChatGPT and Google Bard\label{tfidfimg}}
\end{figure}

Figure \ref{tfidfimg} illustrates an advantage of ChatGPT, with a better average similarity score. Both generative AI tools, ChatGPT and Bard, demonstrate a relatively high percentage of similarity. 

In this method, it is worth noting that both ChatGPT and Bard, our selected generative tools, show a meaningful level of average similarity (image \ref{tab:bigram_analysis_results}), hovering around the 37-47 \% mark. This shows a visible connection between the RWD generated by these models and existing medical research from sources like Pubmed. The \ref{tfidfimg} plot shows that the difference is bigger with a bigger sample of documents. At first, the difference is about 2 \%, but at the end, jumps to about 6 \%. In table \ref{tab:bigram_analysis_results} we see the value of similarity for example 50 documents is 47 \% for ChatGPT and about 37 \% for Google Bard, so the advantage of ChatGPT is almost 10 \%.

\subsection{Bigram Networks Analysis}

\begin{figure}[H]
    \hspace{0.04\textwidth}
    \begin{minipage}{.38\textwidth}
        \includegraphics[width=1\linewidth]{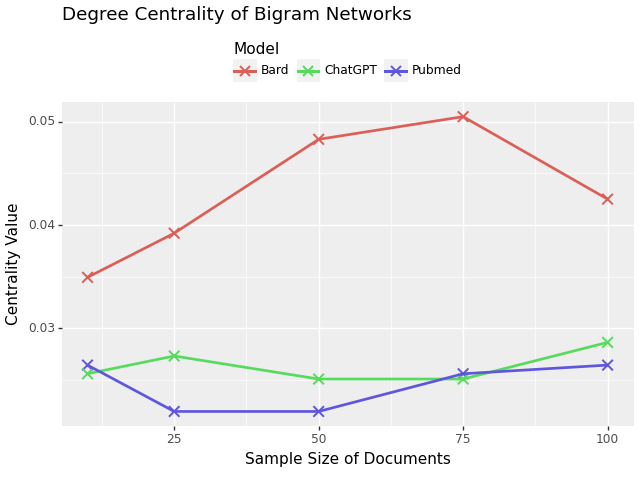}
        \caption{Degree centrality of networks of bigrams from PubMed and generative AI tools documents\label{dcimg}}
    \end{minipage}%
        \hspace{0.12\textwidth}
    \begin{minipage}{.38\textwidth}
        \includegraphics[width=1\linewidth]{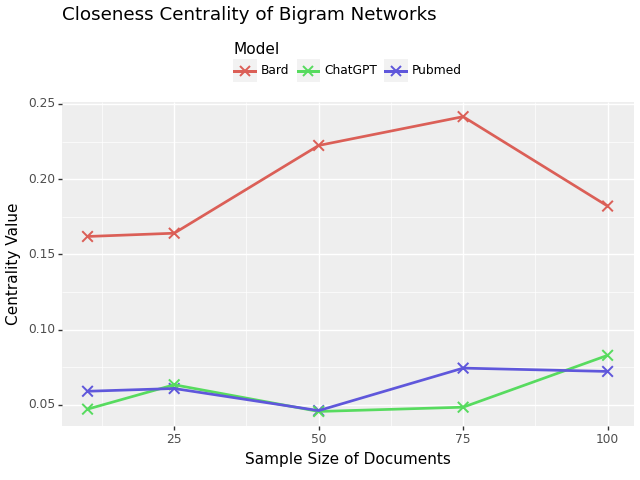}
        \caption{Closeness centrality of networks of bigrams from PubMed and generative AI tools documents\label{ccimg}}
    \end{minipage}
    \hspace{0.04\textwidth}
\end{figure}

\subsubsection{Centrality analysis}
The degree centrality, as seen in plot \ref{dcimg}, displays a structural similarity between the values from PubMed and ChatGPT. In contrast, Bard exhibits much higher values. This suggests that ChatGPT's values are closer to PubMed's and potentially more similar. ChatGPT and PubMed centrality have prevalent degree centrality, between 0.02 and 0.03, while Bard is between 0.035 and 0.05. ChatGPT and Pubmed differences are in the range of 0.001 to 0.005. Differences for Bard and Pubmed are in the range of 0,013 to 0,017. The value for 50 documents in \ref{tab:analysis_results_centrality} is 0.022 and 0.025 for PubMed and ChatGPT networks, it is very close. For Bard, is much higher, about 0.048.

The closeness centrality, in figure \ref{ccimg}, once again, shows a correlation between PubMed and ChatGPT. As seen, Bard has slightly higher values while ChatGPT's values are close to those of PubMed showing that the networks are potentially more similar. PubMed closeness centrality is between 0.05 and 0.08, and for ChatGPT is between 0.05 and 0.89, for Google Bard it is between 0.16 and 0.24. Values are closer to PubMed with a higher number of documents. The differences between ChatGPT and Pubmed are in the range of 0.011 to 0.025, while the differences between Google Bard and Pubmed are in the range of 0.082 to 0.109. For example 50 documents in table \ref{tab:analysis_results_centrality}, the value for PubMed is 0.046, for ChatGPT about 0.045, and for Bard is 0.222. The numbers support ChatGPT's better fit.

\begin{figure}[h]
    \begin{minipage}{.45\textwidth}
        \begin{table}[H]
        \centering
        \caption{Analysis Results: Average Centrality}
        \label{tab:analysis_results_centrality}
        \begin{tabular}{ccccccc}
        \hline
        {\textbf{Dataset size}} & {\textbf{Source}} & \multicolumn{2}{c}{\textbf{Centrality}} \\
        \cline{3-5}
        & & \textbf{Degree} & \textbf{Closeness}\\
        \hline
        \textit{10} & \textit{PubMed} & 0.0264 & 0.0590 & \\
        \cline{2-5}
        & \textit{ChatGPT} & 0.0256 & 0.0471 & \\
        \cline{2-5}
        & \textit{Bard} & 0.0349 & 0.1618 & \\
        \hline
        \textit{25} & \textit{PubMed} & 0.0219 & 0.0608 & \\
        \cline{2-5}
        & \textit{ChatGPT} & 0.0273 & 0.0633 & \\
        \cline{2-5}
        & \textit{Bard} & 0.0392 & 0.1640 & \\
        \hline
        \textit{50} & \textit{PubMed} & 0.0219 & 0.0461 & \\
        \cline{2-5}
        & \textit{ChatGPT} & 0.0250 & 0.0456 & \\
        \cline{2-5}
        & \textit{Bard} & 0.0483 & 0.2223 & \\
        \hline
        \textit{75} & \textit{PubMed} & 0.0256 & 0.0744 & \\
        \cline{2-5}
        & \textit{ChatGPT} & 0.0250 & 0.0484 & \\
        \cline{2-5}
        & \textit{Bard} & 0.0505 & 0.2415 & \\
        \hline
        \textit{100} & \textit{PubMed} & 0.0264 & 0.0722 & \\
        \cline{2-5}
        & \textit{ChatGPT} & 0.0286 & 0.0829 & \\
        \cline{2-5}
        & \textit{Bard} & 0.0425 & 0.1821 & \\
        \hline
        \end{tabular}
        \end{table}
    \end{minipage}%
        \hspace{0.1\textwidth}
    \begin{minipage}{.45\textwidth}
        \begin{table}[H]
            \caption{Bigram ranks in Pubmed and generated datasets}\label{tab:bigram_comparison}
            \begin{tabular}{@{}llll@{}}
                \hline
                Rank & PubMed Bigrams & ChatGPT & Bard\\
                \hline
                1 & prostate cancer & 1 & 1 \\
                2 & radiation therapy & 8 & -- \\
                3 & radical prostatectomy & 28 & -- \\
                4 & localized prostate & 32 & -- \\
                5 & prostate specific & -- & -- \\
                6 & androgen deprivation & -- & -- \\
                7 & cancer patient & 4 & 47 \\
                8 & specific antigen & -- & -- \\
                9 & external beam & -- & -- \\
                10 & free survival & 43 & -- \\
                11 & quality life & 5 & 44 \\
                12 & patient treated & -- & -- \\
                13 & risk prostate & 22 & 6 \\
                14 & deprivation therapy & -- & -- \\
                15 & intermediate risk & -- & -- \\
                16 & intermediate therapy & -- & -- \\
                \hline
            \end{tabular}
        \end{table}
    \end{minipage}
\end{figure}

\section{Discussion}
In this work, we presented a text-mining and network analysis approached to establish the similarity between generated AI contents from tool such as ChatGPT and Google Bard, and biomedical records related to prostate cancer. This enabled this research to be both experimental and quantitative. Therefore, the similarities driven are not subjective and does not suffer human error. 

For the \textbf{text-mining analysis}, we compared the documents on its own entirety to establish on the similarity on the macro level. Using samples of sizes (19, 25, 50, 75, and 100) against the Cosine, Jaccard similarities, we observed that ChatGPT scored a closer similarity than Google Bard. The Cosine similarity for ChatGPT ranged from (35\% - 38\%) , and (32\% - 34\%) for Google Bard. As for the Jaccard Similarity, the similarity ranged from (20\% - 22\%) for ChatGPT, while the similarity ranged from (17\% - 19\%) for Google Bard. This analysis in two different measures favors ChatGPT over Google Bard. We also observed in general that though Jaccard and Cosine may offer scores of different ranges, however, the measures were not fundamentally different from each other, where the Cosine similarity scores are 10\% higher that the similarity scores of the Jaccard. 

\begin{figure}[H]
    \begin{minipage}{.28\textwidth}
        \includegraphics[width=1\linewidth]{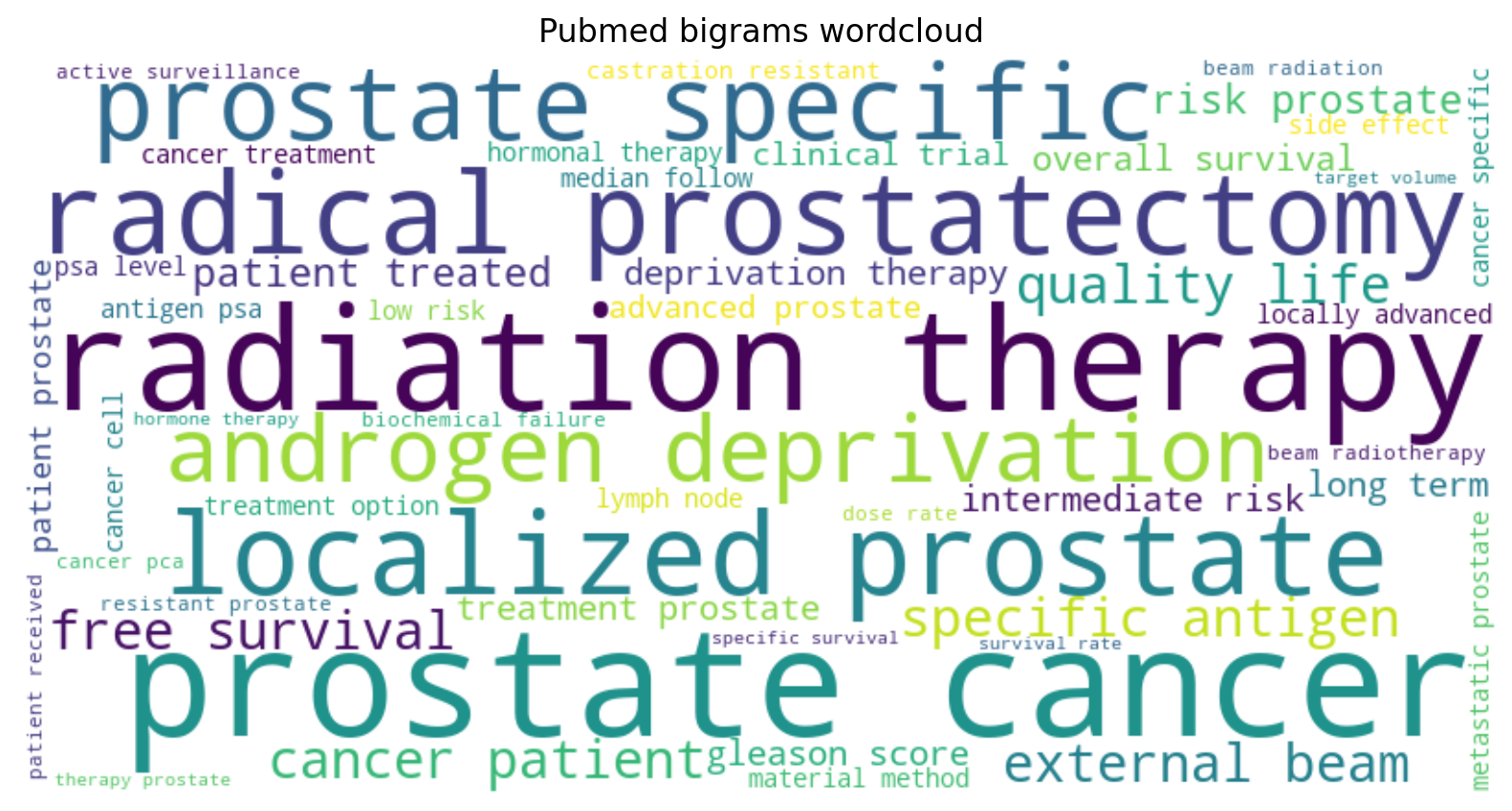}
        \caption{PubMed word cloud}
        \label{fig:pubmedwc}
    \end{minipage}%
        \hspace{0.07\textwidth}
    \begin{minipage}{.28\textwidth}
        \includegraphics[width=1\linewidth]{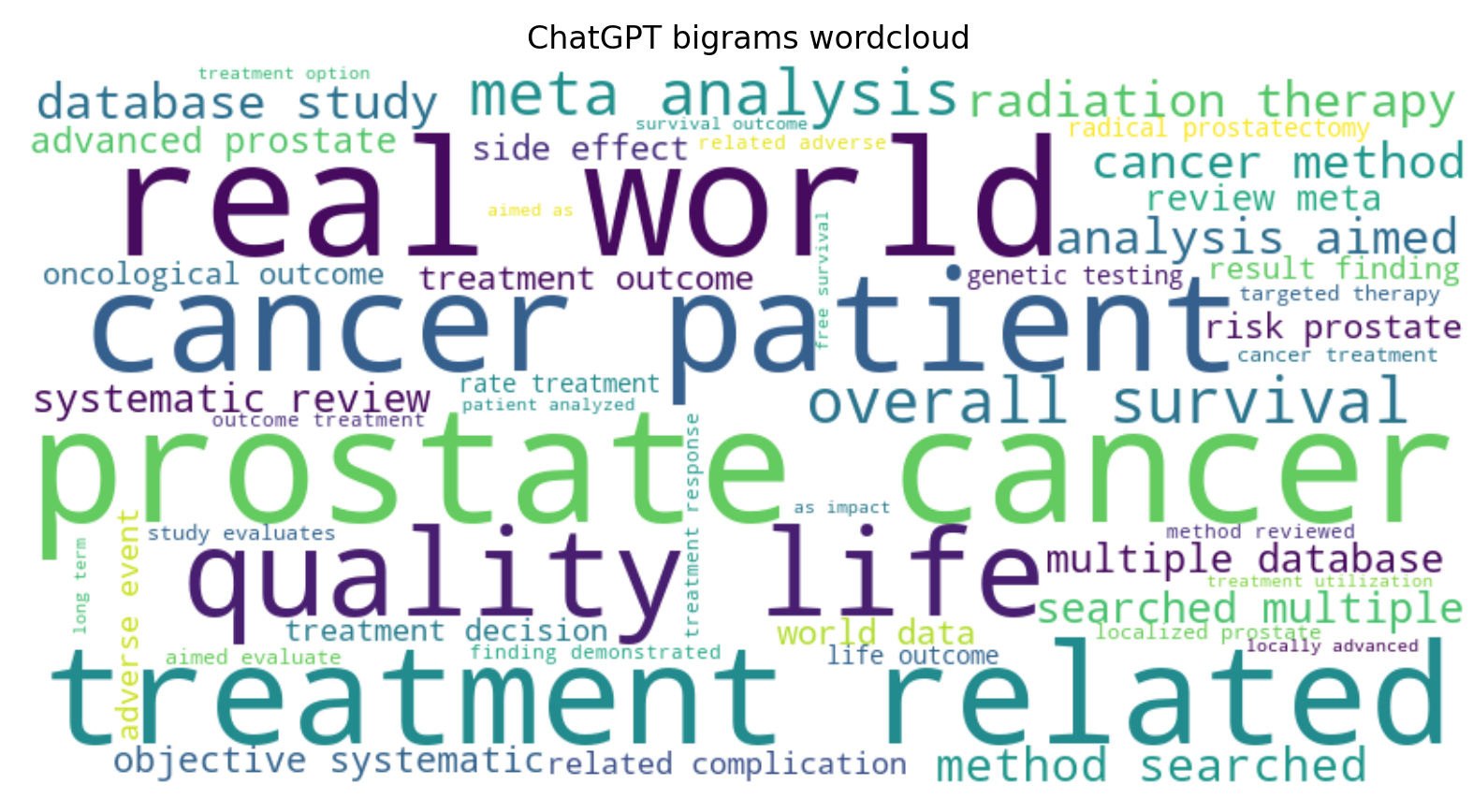}
        \caption{ChatGPT word cloud}
        \label{fig:chatgptwc}
    \end{minipage}%
        \hspace{0.07\textwidth}
    \begin{minipage}{.28\textwidth}
        \includegraphics[width=1\linewidth]{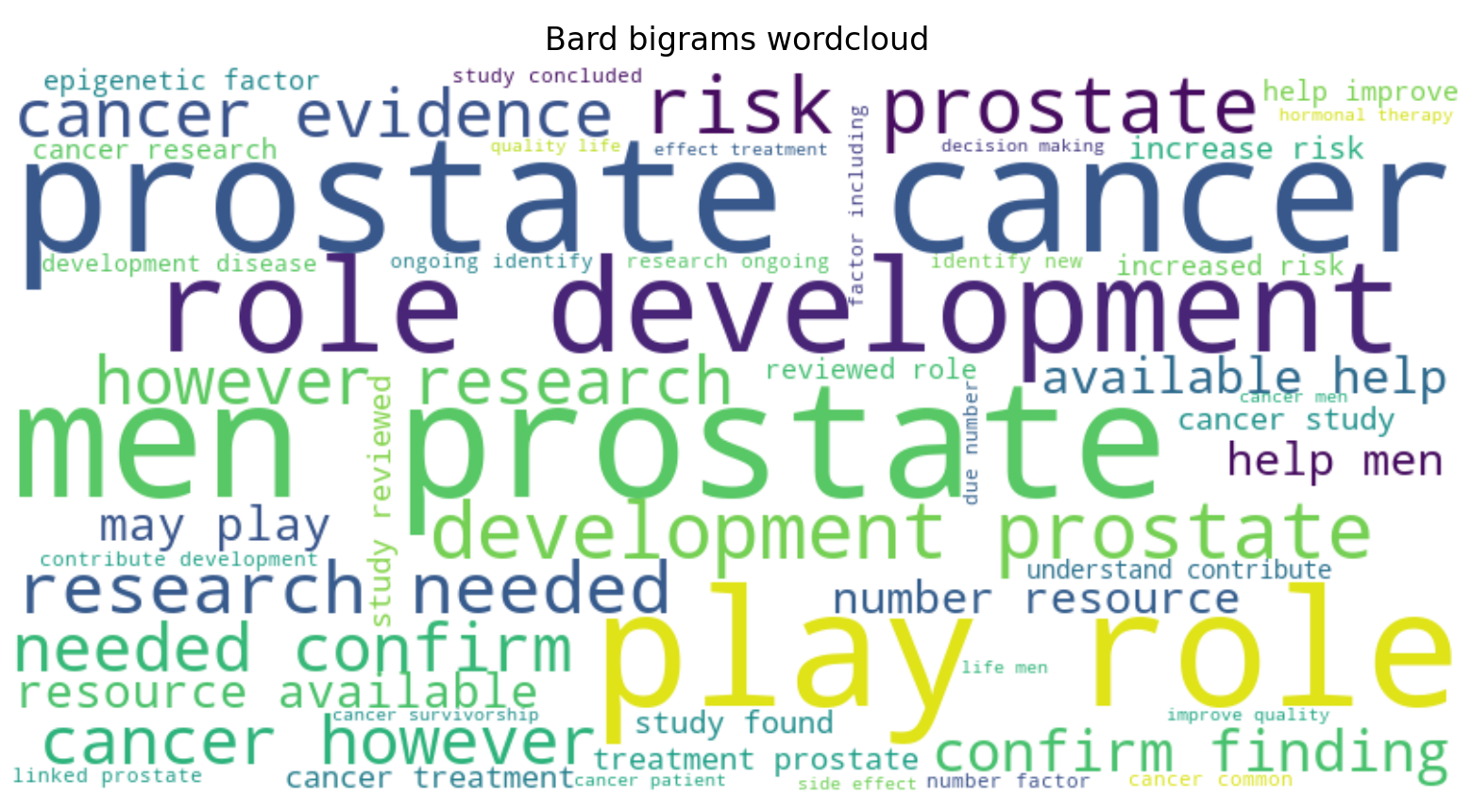}
        \caption{Bard word cloud}
        \label{fig:bardwc}
    \end{minipage}
\end{figure}

From a micro-level, where we analyzed top-k bigrams scores from each of the two sources. We observed a pattern that ChatGPT is consistently beating Google Bard where the difference was widening the bigger the dataset sample. Here we observe the different between the averaged bigram scores: (Size 10: $2\%$, Size 25: $4\%$, Size 50: $9\%$, Size 75: $7\%$, and Size 100: $6\%$). Figures: \ref{fig:pubmedwc}, \ref{fig:chatgptwc}, \ref{fig:bardwc} show word-cloud representation of the top-50 most frequent bigrams. Table \ref{tab:bigram_comparison} shows a snapshot of the top-k (for mere representation purposes) where we observed that ChatGPT offered 8 overlapping bigrams, Google Bard offered 4 bigrams overlapped. While both ChatGPT and Google Bard agreed with PubMed on ``prostate cancer'' as number 1 ranked bigram, they varied on the ranks of the overlapping bigrams. 

We also found an interesting observation in two of the overlapping bigrams: ``cancer patients'' bigram which was ranked as number 7 in the PubMed dataset, it was ranked number 4 in ChatGPT. Also, the ``quality life'' bigram, which was ranked as number 11 in PubMed dataset, it was found to be ranked as number 5 in the ChatGPT dataset. As for Google Bard, ``cancer patient'' was number 47, and ``quality life'' was ranked as number 44. This could be an indication that ChatGPT was trained on dataset that is related to patience wellness that takes quality of life into consideration, while PubMed data is more about the clinical/scientific aspects of the diseases. As for Google Bard, it is not clear why such lower ranking was observed if it is existed. This analysis concludes that ChatGPT is more similar on the macro level (document) and the micro level (bigrams). The full analysis of the bigram ranking can be observed in table \ref{tab:bigram_comparison}. The results also triggers many other questions such as: (1) why the bigrams in generated datasets are different, and why they have different ranks; (2) why does the one bigram exist in ChatGPT and does not exist in Google Bard and much more? These differences in bigrams can help ask various valuable questions.

As presented in the Results section, the \textbf{Network Analysis} section offers us another interesting dimension to compare the similarities. The results of the experiments show that ChatGPT graphs exhibit closer similarities in structure and centrality when compared with PubMed graphs. We observed that the difference between ChatGPT and PubMed networks was about 0.08 for closeness centrality and 0.03 for degree centrality (see Table \ref{tab:analysis_results_centrality}), which is trivial. Google Bard, however, showed a more noticeable and significant difference of 0.2 and 0.05, respectively (as shown in Table \ref{tab:analysis_results_centrality}).

Though the analysis here shows that ChatGPT achieved better results using various similarity measures, we state that these results are within to the scope of the ``prostate cancer'' dataset from the multiple sources and bound to the version tools used. For ChatGPT, we used version 3.5 (not 4.0), and for Google Bard, we used the version before the launch of Google Gemini, which now empowers Google Bard. Therefore, we believe that for our study to be conclusive, it is necessary to perform further analysis on various datasets from multiple diseases (e.g., diabetes, depression, cardiovascular, among others). It will also be necessary to use the most recent APIs to have a more conclusive analysis.

\section{Future Directions}
We recently aquired access to ChatGPT AIPs (version 3.5), therefore, gathering new datasets related to various diseases is one of the immediate directions to analyze and conclude the study here. When more conclusive analysis is performed, we will investigate whether any of these tools offer real world evidence when proven authentic and complimentary. It is our hope that we can use such tools to ask interesting question that lead researchers to investigate better hypothesis and make significant findings.  

\section{Acknowledgments}
This publication is partially supported by the European Union’s Horizon 2020 research and innovation programme under grant agreement Sano No. 857533 and carried out within the International Research Agendas programme of the Foundation for Polish Science, co-financed by the European Union under the European Regional Development Fund. Additionally is partially created as part of the Ministry of Science and Higher Education's initiative to support the activities of Excellence Centers established in Poland under the Horizon 2020 program based on the agreement No MEiN/2023/DIR/3796. 

\bibliographystyle{splncs04}
\bibliography{reference}

\begin{thebibliography}{10}
\providecommand{\url}[1]{\texttt{#1}}
\providecommand{\urlprefix}{URL }
\providecommand{\doi}[1]{https://doi.org/#1}

\bibitem{bard}
Google bard. \url{https://bard.google.com/}, accessed: 2023-08-03

\bibitem{ngramint}
An introduction to n-grams: What are they and why do we need them?,
  \url{https://blog.xrds.acm.org/2017/10/introduction-n-grams-need/}

\bibitem{pubmed}
National library of medicine - pubmed. \url{https://pubmed.ncbi.nlm.nih.gov/},
  accessed: 2023-08-03

\bibitem{chatgpt}
Openai chatgpt. \url{https://chat.openai.com/}, accessed: 2023-08-03

\bibitem{Liu2023}
Using ai-generated suggestions from chatgpt to optimize clinical decision
  support. Journal of the American Medical Informatics Association
  \textbf{30},  1237--1245 (6 2023). \doi{10.1093/jamia/ocad072}

\bibitem{abdeen2021fighting}
Abdeen, M.A., Hamed, A.A., Wu, X.: Fighting the covid-19 infodemic in news
  articles and false publications: The neonet text classifier, a supervised
  machine learning algorithm. Applied Sciences  \textbf{11}(16), ~7265 (2021)

\bibitem{Agarwal2014}
Agarwal, N., Rawat, M., Vijay, M.: Comparative analysis of jaccard coefficient
  and cosine similarity for web document similarity measure. International
  Journal for Advance Research in Engineering and Technology  \textbf{2},
  18--21 (2014)

\bibitem{Agossah2023}
Agossah, A., Krupa, F., Silva, M.P.D., Callet, P.L.: Llm-based interaction for
  content generation: A case study on the perception of employees in an it
  department. pp. 237--241. Association for Computing Machinery (ACM) (6 2023).
  \doi{10.1145/3573381.3603362}

\bibitem{bard3}
AI, G.: Pathways language model (palm): Scaling to 540 billion parameters for
  breakthrough performance (2022),
  \url{ai.googleblog.com/2022/04/pathways-language-model-palm-scaling-to.html}

\bibitem{Andre2022}
Andre, E.B., Carrington, N., Siami, F.S., Hiatt, J.C., McWilliams, C., Hiller,
  C., Surinach, A., Zamorano, A., Pashos, C.L., Schulz, W.L.: The current
  landscape and emerging applications for real-world data in diagnostics and
  clinical decision support and its impact on regulatory decision making (12
  2022). \doi{10.1002/cpt.2565}

\bibitem{pmlrv70}
Bach, S.H., He, B., Ratner, A., R{\'e}, C.: Learning the structure of
  generative models without labeled data. In: Precup, D., Teh, Y.W. (eds.)
  Proceedings of the 34th International Conference on Machine Learning.
  Proceedings of Machine Learning Research, vol.~70, pp. 273--282. PMLR (06--11
  Aug 2017), \url{https://proceedings.mlr.press/v70/bach17a.html}

\bibitem{baumfeld2022current}
Baumfeld~Andre, E., Carrington, N., Siami, F.S., Hiatt, J.C., McWilliams, C.,
  Hiller, C., Surinach, A., Zamorano, A., Pashos, C.L., Schulz, W.L.: The
  current landscape and emerging applications for real-world data in
  diagnostics and clinical decision support and its impact on regulatory
  decision making. Clinical Pharmacology \& Therapeutics  \textbf{112}(6),
  1172--1182 (2022)

\bibitem{Chung2023}
Chung, J., Kamar, E., Amershi, S.: Increasing diversity while maintaining
  accuracy: Text data generation with large language models and human
  interventions. pp. 575--593. Association for Computational Linguistics (ACL)
  (8 2023). \doi{10.18653/v1/2023.acl-long.34}

\bibitem{Guo2023}
Guo, J., Li, J., Li, D., Tiong, A.M.H., Li, B., Tao, D., Hoi, S.: From images
  to textual prompts: Zero-shot visual question answering with frozen large
  language models. pp. 10867--10877. Institute of Electrical and Electronics
  Engineers (IEEE) (8 2023). \doi{10.1109/cvpr52729.2023.01046}

\bibitem{hamed2015measuring}
Hamed, A.A., Ayer, A.A., Clark, E.M., Irons, E.A., Taylor, G.T., Zia, A.:
  Measuring climate change on twitter using google’s algorithm: Perception
  and events. International Journal of Web Information Systems  \textbf{11}(4),
   527--544 (2015)

\bibitem{hamed2023challenging}
Hamed, A.A., Lee, B.S., Crimi, A., Misiak, M.M.: Challenging the machinery of
  generative ai with fact-checking: Ontology-driven biological graphs for
  verifying human disease-gene links (2023)

\bibitem{hamed2023improving}
Hamed, A.A., Wu, X.: Improving detection of chatgpt-generated fake science
  using real publication text: Introducing xfakebibs a supervised-learning
  network algorithm (2023)

\bibitem{Kao2023}
Kao, H.J., Chien, T.W., Wang, W.C., Chou, W., Chow, J.C.: Assessing chatgpt's
  capacity for clinical decision support in pediatrics: A comparative study
  with pediatricians using kidmap of rasch analysis (6 2023).
  \doi{10.1097/MD.0000000000034068}

\bibitem{Kim2019}
Kim, S.W., Gil, J.M.: Research paper classification systems based on tf-idf and
  lda schemes. Human-centric Computing and Information Sciences  \textbf{9} (12
  2019). \doi{10.1186/s13673-019-0192-7}

\bibitem{Koraishi2023}
Koraishi, O.: Teaching english in the age of ai: Embracing chatgpt to optimize
  efl materials and assessment. Language Education and Technology  \textbf{3},
  55--72 (2023), \url{https://bit.ly/43en7e1}

\bibitem{Lee2023a}
Lee, A.: What are large language models used for and why are they important?
  (2023),
  \url{https://blogs.nvidia.com/blog/2023/01/26/what-are-large-language-models-used-for/}

\bibitem{Liao2023}
Liao, Z., Wang, J., Shi, Z., Lu, L., Tabata, H.: Revolutionary potential of
  chatgpt in constructing intelligent clinical decision support systems (2023).
  \doi{10.1007/s10439-023-03288-w}

\bibitem{Liu2023b}
Liu, J., Wang, C., Liu, S.: Utility of chatgpt in clinical practice. Journal of
  Medical Internet Research  \textbf{25} (2023). \doi{10.2196/48568}

\bibitem{Liu2019a}
Liu, S., Demirel, M.F., Liang, Y.: N-gram graph: Simple unsupervised
  representation for graphs, with applications to molecules. vol.~32 (2019)

\bibitem{Meng2021}
Meng, X., Li, W., Peng, X., Li, Y., Li, M.: Protein interaction networks:
  centrality, modularity, dynamics, and applications (2021).
  \doi{10.1007/s11704-020-8179-0}

\bibitem{Moro2023}
Moro, A., Greco, M., Cappa, S.F.: Large languages, impossible languages and
  human brains. Cortex  \textbf{167},  82--85 (10 2023).
  \doi{10.1016/j.cortex.2023.07.003}

\bibitem{Mu2023}
Mu, Y., Reheman, A., Cao, Z., Fan, Y., Li, B., Li, Y., Xiao, T., Zhang, C.,
  Zhu, J.: Augmenting large language model translators via translation
  memories. pp. 10287--10299. Association for Computational Linguistics (ACL)
  (8 2023). \doi{10.18653/v1/2023.findings-acl.653}

\bibitem{Roupen2024}
Odabashian, R., Bastin, D., Jones, G., Manzoor, M., Tangestaniapour, S., Assad,
  M., Lakhani, S., Odabashian, M., McGee, S.: Assessment of chatgpt-3.5's
  knowledge in oncology: Comparative study with asco-sep benchmarks. JMIR AI
  \textbf{3},  e50442 (Jan 2024). \doi{10.2196/50442},
  \url{https://ai.jmir.org/2024/1/e50442}

\bibitem{Orchard2023}
Orchard, T., Tasiemski, L.: The rise of generative ai and possible effects on
  the economy. Economics and Business Review  \textbf{9},  9--26 (4 2023).
  \doi{10.18559/ebr.2023.2.732}

\bibitem{Rani2021}
Rani, U., Bidhan, K.: Comparative assessment of extractive summarization:
  Textrank, tf-idf and lda. Journal of scientific research  \textbf{65},
  304--311 (2021). \doi{10.37398/jsr.2021.650140}

\bibitem{Rao2023}
Rao, A., Pang, M., Kim, J., Kamineni, M., Lie, W., Prasad, A.K., Landman, A.,
  Dreyer, K.J., Succi, M.D.: Assessing the utility of chatgpt throughout the
  entire clinical workflow. medRxiv : the preprint server for health sciences
  (2 2023). \doi{10.1101/2023.02.21.23285886},
  \url{http://www.ncbi.nlm.nih.gov/pubmed/36865204}

\bibitem{tfidf}
Sammut, C., Webb, G.I. (eds.): TF--IDF, pp. 986--987. Springer US, Boston, MA
  (2010)

\bibitem{Schad2022}
Schad, F., Thronicke, A.: Real-world evidence—current developments and
  perspectives (2022). \doi{10.3390/ijerph191610159}

\bibitem{Shortliffe}
Shortliffe, E.H.: Role of evaluation throughout the life cycle of biomedical
  and health ai applications. BMJ health \& care informatics  \textbf{30}(1),
  e100925 (December 2023). \doi{10.1136/bmjhci-2023-100925}

\bibitem{bard2}
Siad, S.M.: The promise and perils of google's bard for scientific research. AI
   (03 2023). \doi{10.17613/yb4n-mc79}

\bibitem{smithaiblog}
{Smith.ai}: Where do generative ai models source their data \& information?
  (Year of Publication),
  \url{https://smith.ai/blog/where-do-generative-ai-models-source-their-data-information},
  accessed on October 27, 2023

\bibitem{Thada2013}
Thada, V., Jaglan, V.: Comparison of jaccard, dice, cosine similarity
  coefficient to find best fitness value for web retrieved documents using
  genetic algorithm. International Journal of Innovations in Engineering and
  Technology  \textbf{2},  202--205 (2013),
  \url{http://www.dknmu.org/uploads/file/6842.pdf}

\bibitem{touvron2023llama}
Touvron, H., Lavril, T., Izacard, G., Martinet, X., Lachaux, M.A., Lacroix, T.,
  Rozière, B., Goyal, N., Hambro, E., Azhar, F., Rodriguez, A., Joulin, A.,
  Grave, E., Lample, G.: Llama: Open and efficient foundation language models
  (2023)

\bibitem{fdadocument}
{U.S. Food and Drug Administration}: Framework for fda’s real-world evidence
  program (Year of Publication),
  \url{https://www.fda.gov/media/120060/download}, accessed on October 27, 2023

\bibitem{Wang2008}
Wang, G., Shen, Y., Luan, E.: Measure of centrality based on modularity matrix.
  Progress in Natural Science  \textbf{18} (2008).
  \doi{10.1016/j.pnsc.2008.03.015}

\bibitem{Wang2019}
Wang, J., Xu, W., Yan, W., Li, C.: Text similarity calculation method based on
  hybrid model of lda and tf-idf. pp.~1--8. Association for Computing Machinery
  (12 2019). \doi{10.1145/3374587.3374590}

\bibitem{Zhang2017}
Zhang, J., Luo, Y.: Degree centrality, betweenness centrality, and closeness
  centrality in social network. Atlantis Press (2017).
  \doi{10.2991/msam-17.2017.68}

\end{thebibliography}
%




\end{document}